\documentclass[conference]{IEEEtran}

\usepackage{graphicx}
\usepackage{amsmath, amssymb}
\usepackage{cite}
\usepackage{caption}
\usepackage{booktabs}
\usepackage{float}
\usepackage{algorithm}
\usepackage{algorithmic}
\usepackage{placeins} 

\begin{document}

\title{Generating Hierarchical JSON Representations of Scientific Sentences Using LLMs}

\author{
\IEEEauthorblockN{Satya Sri Rajiteswari Nimmagadda}
\IEEEauthorblockA{Marshall University\\
Email: nimmagadda2@marshall.edu}
\and
\IEEEauthorblockN{Ethan Young}
\IEEEauthorblockA{Marshall University\\
Email: young537@marshall.edu}
\and
\IEEEauthorblockN{Niladri Sengupta}
\IEEEauthorblockA{Fractal Analytics Inc\\
Email: dinophysicsiitb@gmail.com}
\and
\IEEEauthorblockN{Ananya Jana}
\IEEEauthorblockA{Marshall University\\
Email: jana@marshall.edu}
\and
\IEEEauthorblockN{Aniruddha Maiti}
\IEEEauthorblockA{West Virginia State University\\
Email: aniruddha.maiti@wvstateu.edu}
}

\maketitle

\begin{abstract}
This paper investigates whether structured representations can preserve the meaning of scientific sentences. To test this, a lightweight LLM is fine-tuned using a novel structural loss function to generate hierarchical JSON structures from sentences collected from scientific articles. These JSONs are then used by a generative model to reconstruct the original text. Comparing the original and reconstructed sentences using semantic and lexical similarity we show that hierarchical formats are capable of retaining information of scientific texts effectively.

\end{abstract}


\section{Introduction}
Scientific sentences often express a core claim along with additional information such as conditions, temporal qualifiers, or logical exceptions. Any representation used for processing scientific text must reflect this complex structure to avoid information loss. Previous methods mapped sentences to a single relation between two entities using a flat schema such as \textit{A–relation–B} \cite{maiti2025comparative}. While suitable for simple assertions, this format cannot represent clause-level dependencies or hierarchical logic. This study uses a hierarchical JSON format that separates the core statement from supporting components. A lightweight language model is fine-tuned to generate such structures from raw sentences. The structured JSON output is then used by a generative model to reconstruct the original sentence. Reconstruction quality is measured using semantic and lexical similarity to assess whether the structured format generated by LLM preserved meaning. The method is evaluated on a diverse set of scientific sentences sampled from multiple scientific domains.

\section{Background and Related Work}
Several studies examined sentence reconstruction from latent spaces \cite{song2020information, seputis2025rethinking}. Maiti et al. represented scientific sentences to a flat \textit{A--relation--B} structure \cite{maiti2025comparative}. Wang and Bai~\cite{wang2024entity} represented long sentences as dependency graphs and used graph network to capture grammatical and semantic relations. Studies on structured representation learning include work by Liu et al.~\cite{liu2025sstag},  Zheng et al.~\cite{zheng2022double}. Das et al.~\cite{das2025hypersumm} and Thilagam et al.~\cite{thilagam2023leveraging} used JSON structures in their work. Yang et al.~\cite{yang2025structeval} introduced a benchmark for structured outputs in formats such as JSON, HTML, and YAML. Recent studies examined structured output from LLMs. Shorten et al.~\cite{shorten2024structuredrag} reported variation in JSON formatting behavior in LLM responses. Dunn et al.~\cite{dunn2022structured} fine tuned GPT 3 for structured extraction from scientific text. Dagdelen et al.~\cite{dagdelen2024structured} extended this type of extraction to scientific records in JSON format.

\section{Methodology}
\subsection{Dataset}
We constructed a dataset of 1500 scientific sentences taken from articles in seven scientific domains: physics, computer science, mathematics, economics, biology, chemistry, and medicine. Articles were chosen from arXiv, bioRxiv, ChemRxiv, and PubMed. Sentences with equations, symbols, or citation markers were excluded in the preprocessing step. After preprocessing, out of 1500 initial sentences, 1370 having complete meaning were retained. Table \ref{table:data_statistics} reports the number of articles and sentences collected from each source. No more than six sentences were taken from one article.

\begin{table}[h!]
\centering
\begin{tabular}{|l|l|c|r|}
\hline
\textbf{Domain} & \textbf{Repository} & \textbf{Articles} & \textbf{Sentences} \\
\hline
Physics          & arXiv     & 97  & 447  \\
Computer Science & arXiv     & 39  & 193  \\
Mathematics      & arXiv     & 25  & 113  \\
Economics        & arXiv     & 50  & 249  \\
Biology          & bioRxiv   & 51  & 256  \\
Chemistry        & ChemRxiv  & 12  & 58   \\
Medicine         & PubMed    & 17  & 54   \\
\hline
\textbf{Total}   & \textbf{4 (unique)} & \textbf{291} & \textbf{1370} \\
\hline
\end{tabular}
\caption{Dataset summary by domain and repository.}
\label{table:data_statistics}
\end{table}

For experiments, we divided the collected and cleaned data into 1096 training and validation (n=138) sentences, and 274 test sentences.
\subsection{Generation of Training Data}
GPT-4o is used to generate high-quality training data. The prompt contained instruction to generate output JSONs from input sentences containing two main parts:

\noindent \textbf{\texttt{core: }} This part has two keys: a generic label, and a central claim of the sentence so that the most important information of the sentence can be retained.
\textbf{\texttt{hierarchy: }} This part contained the hierarchical structure around the core. Each level of the hierarchy contains two important keys: a relationship type and components connected by that relationship type. The generative model is instructed to choose relation types freely, but 17 predefined relationship types were given as a reference or example. These relationship types are a refined list used in prior work \cite{maiti2025comparative}.

The JSON generating prompt contained instructions to: (a) avoid word-to-word copying from the input; (b) compress the contents so that no single field (either in core or any hierarchy) exceeds 30\% of the original sentence length; (c) organize content hierarchically as per relationship types based on logical flow of the sentence. In the prompt, we also provided few shot examples showing sentences and their ideal JSON representations. 

\subsection{Fine Tuning of LLM}
\label{subsec:fine-tuneLLM}
We selected Mistral-7B model \cite{ahmad2025multi}, to demonstrate and validate the idea. The selected model is a representative of $\sim$7B parameter LLM class. The model is fine-tuned to predict the structured output sequence from the input sentences using low-rank adaptation (LoRA) technique. 

\begin{algorithm}[H]
\caption{Novel Structural Loss Function used in this study}
\label{alg:json_loss}
\begin{algorithmic}[1]
\REQUIRE Generated token logits from causal language model
\STATE Compute base loss: $\mathcal{L}_{\text{CE}} \leftarrow$ CrossEntropyLoss
\STATE Decode predictions: $\hat{y} \leftarrow \text{ArgMax}(\text{logits})$
\STATE Initialize JSON failure count: $f \leftarrow 0$
\FOR{each decoded string $s$ in $\hat{y}$}
    \STATE \# Attempt to extract structured output: 
    \STATE $j \leftarrow \text{JSONParser}(s)$
    \IF{JSON parsing fails}
        \STATE $f \leftarrow f + 1$
    \ENDIF
\ENDFOR
\STATE Compute structure penalty: $\mathcal{L}_{\text{struct}} \leftarrow f / |\hat{y}|$
\STATE Combine losses: $\mathcal{L}_{\text{total}} \leftarrow \mathcal{L}_{\text{CE}} + \mathcal{L}_{\text{struct}}$
\RETURN $\mathcal{L}_{\text{total}}$
\end{algorithmic}
\end{algorithm}

To enforce syntactic correctness in structured outputs, we used a novel training mechanism that integrates a runtime structure validation agent into the training loop. Along with token-level losses, the proposed method also used penalty by dynamically evaluating whether the generated output is a valid JSON object. This is achieved by decoding each generated sample and applying an external parser (along with a Python \textit{try–except} block) to check JSON validity. If parsing fails, a penalty is applied to the batch loss. We described this process in Algorithm~\ref {alg:json_loss}.

The trained model was then used to generate structured JSON representations of the held-out test set.

\subsection{Sentence Reconstruction}
\label{subsec:sengt_reconstruct}
To assess whether JSONs generated from the test data by Mistral-7B model retained sufficient information to reconstruct the original scientific sentences, they were used as input to a generative model (GPT-4o) API. We created a prompt that instructs the API to produce a natural language reconstruction of the given JSON. The generative model was tasked to interpret the core claim, hierarchical components and relation types and generated one coherent sentence per input.

\section{Experiments}
\subsection{Training Configuration and Model Details}
We fine-tuned the \emph{Mistral-7B-v0.1} model using the training and validation data. LoRA adapters were applied to attention and feedforward projection layers, the rest of the model was kept frozen. The LoRA ~\cite{hu2022lora} configuration used rank 16, scaling factor 16, and dropout of 0.05. Training was performed using fully sharded data parallelism with automatic layer wrapping. We enabled gradient checkpointing to reduce memory use. We also used bfloat16 precision to reduce memory footprints. The model was trained for five epochs. The per-device batch size was 1, with gradient accumulation over 4 steps. The learning rate was set to $2 \times 10^{-4}$ with a cosine decay schedule and a warmup ratio of 0.1. The maximum gradient norm was clipped to 0.3. Input length was capped at 2048 tokens. We applied proposed loss penalty as described in Algorithm-\ref{alg:json_loss}. 

\subsection{Evaluation}
We assessed the information retention capability of JSON structures by comparing the original and reconstructed sentences (refer to Subsection-\ref{subsec:sengt_reconstruct}) using semantic similarity and lexical overlap. Cosine similarity between sentence embeddings of the original and reconstructed sentences is used to measure semantic similarity. Sentence transformer, \emph{all-mpnet-base-v2}, is used to convert sentences into embedding vectors.
BLEU\cite{papineni2002bleu}, ROUGE 1 F1\cite{lin2004rouge}, and METEOR \cite{banerjee2005meteor} are used for measuring lexical overlap. 

\section{Results}
All JSON outputs (100\%) from test data (n=274) were of valid JSON format. This indicates, irrespective of the content of output JSON, the model learned to produce structurally correct JSON format.
The reconstruction stage generated one sentence from each JSON. The aggregated comparison result with the original sentences is shown in Table \ref{tab:test_summary_stats}.
\begin{table}[H]
\centering
\caption{Summary Statistics on Test Set (n = 274)}
\label{tab:test_summary_stats}
\begin{tabular}{lcc}
\toprule
\textbf{Metric} & \textbf{Mean} & \textbf{Std Dev} \\
\midrule
Cosine Similarity & 0.8724 & 0.0682 \\
BLEU & 0.1496 & 0.1215 \\
ROUGE-1 F1 & 0.5725 & 0.1138 \\
METEOR & 0.4907 & 0.1405 \\
\bottomrule
\end{tabular}
\end{table}
The cosine similarity values remain above 0.85 for most sentences, which proves consistent meaning retention. 

\begin{table}[H]
\centering
\caption{Example Reconstructed and Original Sentences}
\label{tab:example_reconstruction}
\begin{tabular}{l p{3.6cm} p{3.6cm}}
\toprule
\textbf{ID} & \textbf{Original} & \textbf{Reconstructed} \\
\midrule
97 & Nonlocal boxes, theoretical objects that violate a bipartite Bell inequality as much as the no-signaling principle allows and which are physically impossible for most scenarios, are feasible if the two parties have 3 measurements with 4 outputs.  & Nonlocal boxes are feasible with specific measurements, given the condition of 3 measurements and 4 outputs, although they theoretically violate the bipartite Bell inequality and the no-signaling principle, making them physically impossible in most scenarios. \\

185 & The study investigates the electronic properties of complex materials using advanced computational methods. & The study investigates the electronic properties of materials using advanced computational methods. \\

134 & Our formulation and solution framework are to attain Pareto-optimal joint BS association and beamforming design strategies with guaranteed SINRS at the MSS. & The network design strategy involves the formulation and solution for optimal design, which is different from previous works, aiming for pareto-optimal strategies and guaranteed SINRS at MSS. \\

522 & Therefore, in several cases they perform poorly compared to the fully digital approaches since they can not exploit the inherent spatial multiplexing gain of the channel. & In the comparison of digital vs. \\
\bottomrule
\end{tabular}
\end{table}

Tables \ref{tab:example_reconstruction}, and \ref{tab:evaluation_scores} show few examples and corresponding evaluation scores. Sentence 185 and 97 have high cosine similarity. Both reconstructions retain the main idea with only minor changes. Although Sentence 185 has good lexical overlap, sentence 97 has moderate lexical overlap between the original and reconstructed versions. The reconstruction presents the scientific context in a slightly different order with some different words in the latter case. Sentence 134 and sentence 522 have low cosine similarity. The reconstruction does not retain the main idea here with minimal lexical overlap.

\begin{table}[H]
\centering
\caption{Evaluation Scores for Example Sentences}
\label{tab:evaluation_scores}
\begin{tabular}{lcccc}
\toprule
\textbf{ID} & \textbf{Cosine} & \textbf{BLEU} & \textbf{METEOR} & \textbf{ROUGE} \\
\midrule
97  & 0.889 & 0.043 & 0.502 & 0.593 \\
185 & 0.944 & 0.788 & 0.934 & 0.960 \\
134 & 0.568 & 0.071 & 0.460 & 0.500 \\
522 & 0.543 & 0.002 & 0.097 & 0.250 \\
\bottomrule
\end{tabular}
\end{table}

\noindent \textbf{Overall Assessment:} The JSON representation retains the meaning of most sentences. The mean cosine similarity value above 0.85 supports this observation. In some cases, we found lower scores which is due to compressed clauses or incomplete logical content. 

\section{Conclusion}
Lightweight LLM models can be finetuned to generate hierarchical JSON representations of scientific sentences. The representation can preserve the meaning in most of the cases as shown in this study by reconstructing the sentences from such JSONs.  Future work will test robustness under corrupted structures and explore tree-based loss for better alignment of semantic roles.

\section*{Acknowledgments}
This research is partially supported by the NASA EPSCoR, \textbf{Grant \#80NSSC22M0027}, through the Seed Grant awarded to Aniruddha Maiti by the NASA WV EPSCoR Committee. \noindent Part of the computational requirement for this work is done using resources at the \textit{Pittsburgh Supercomputing Center (PSC)} and \textit{Anvil at Purdue University} through allocation \textbf{Grant number: CIS250884 (Campus Champion)} of the \textit{Advanced Cyberinfrastructure Coordination Ecosystem: Services and Support (ACCESS)} program.

\bibliographystyle{IEEEtran}
\bibliography{references}

\end{document}